# Decision Under Uncertainty in Diagnosis

Charles I. Kalme
Inference Corporation
5300 West Century Blvd.
Los Angeles, Ca. 90045

**Abstract** This paper describes the incorporation of uncertainty in diagnostic reasoning based on the set covering model of Reggia et. al. extended to what in the Artificial Intelligence dichotomy between deep and compiled (shallow, surface) knowledge based diagnosis may be viewed as the generic form at the compiled end of the spectrum. A major undercurrent in this is advocating the need for a strong underlying model and an integrated set of support tools for carrying such a model in order to deal with uncertainty.

## 1. Introduction

This paper describes the incorporation of uncertainty in diagnostic reasoning based on the set covering model of Reggia et. al. [9] extended to what in the Artificial Intelligence (AI) dichotomy between deep and compiled (shallow, surface) knowledge based diagnosis [2] may be viewed as the generic form at the compiled end of the spectrum. After a description of the underlying model in the main body of the paper, we summarize our position on the choice of uncertainty models, the propagation of uncertainty in networks, and the acquisition of uncertainty values as the major areas in need of development for the success of our implementation. We close out the paper with a selective set of references which becomes comprehensive if the references therein are followed up, particularly [3], which cuts across the entire range of AI and uncertainty from both points of view, AI and uncertainty, and is closest in spirit to our approach.

A major undercurrent in our approach is advocating the need for a strong underlying model and an integrated set of support tools for carrying such a model in order to deal with uncertainty, which we are in the process of developing. This leads to a somewhat nonstandard view in AI in that we strongly advocate the need for incorporating algorithmic knowledge, which has been a theme in all our work [4] and translates here to the incorporation of standard statistics tools as part of our system, very much in line with [3].

## 2. The Underlying Model

Our point of departure is a Fault/Symptom Relation Graph that assigns to each fault (disorder) the set of symptoms (manifestations) associated with it, and diagnosis consists of matching a set of input symptoms to a (set of) fault(s) that best explains the symptoms. The basic diagnostic strategy is a sequencial decision procedure, whereby a set of symptoms is acquired sequentially and matched against possible explanations, which may be faults or sets of faults, depending on whether a single or multiple fault hypothesis is in effect. This continues until a satisfactory explanation is found, or until no further information can be gathered. This may also be viewed as a hypothesize-and-test paradigm in that generally new symptoms or their absence are discovered by querying on the basis of trying to strengthen the conviction concerning some explanation in the suspect set, treated as a hypothesis.

Both the Fault/Symptom Relation Graph and the associated decision procedure may carry considerably more structure. The top level data structure is a Fault/Symptom Relation Graph, which we model as a subset of the product space $C \subseteq F \times S$, where $F$ is the set of all faults, $S$ is the set of all symptoms, and $C$ is the causal relation connecting $F$ to $S$. This graph may be represented implicitly in rules or explicitly in a schema structure, which the authors of the set covering model distinguish as deductive and inductive representations [5], respectively. The individual sets $F$ and $S$ may also have internal structure based on hierarchical classification, including causal relations, and every connection, including the relation $C$, may have uncertainty associated with it.

## 2.1. Deductive versus Abductive Representation

Successful diagnosis consists of deducing from a set of symptoms a fault that explains the symptoms. This suggests representing knowledge in rules of the form **Set of Symptoms ==> Fault**, which is the underlying representation in MYCIN [1]. The authors of the set covering model [5] call this representation deductive, and observe that in practice the knowledge is actually obtained in the form **Fault ==> Associated Set of Symptoms**, which they call abductive, based of the associated hypothesize-and-test diagnostic strategy suggestive of abductive reasoning [7]. They also argue that in general it is not straightforward to invert this relation directly, as required for the deductive form, and that therefore the abductive form is more general.

The authors point out that the distinction carries implications with respect to both data representation and processing, in that in the deductive representation the fault/symptom relation graph is most naturally stored in rules, with an associated rule-based processing paradigm, as in MYCIN [1], whereas in the abductive representation it is more natural to set up a schema representation, one schema for each fault $f$ carrying an attribute slot for each associated symptom $s$, with an associated procedural hypothesize-and-test processing paradigm for inverting this relation, as in their set covering model.

We are in full agreement with the authors up to the point where they shun the rule-based approach altogether. In our opinion not only do experts in their limited domains often manage to invert the relation $C$ directly, but also starting from the abductive form the relation may be inverted algorithmically to yield automated rule generation by decision table methods. Indeed, the associated algorithms constitute a major part of the knowledge acquisition component of our system and are necessary for a viable incorporation of uncertainty. Thus we treat the sets $F$ and $S$ symmetrically, assuming each to have its own schema representation, with both $C$ and its inverse $C^{-1} \subseteq S \times F$, included in the representation, as well as whatever additional classification and uncertainty structure we impose.

## 2.2. Hierarchical Classification

In practice the Fault/Symptom Relation Graph may have considerably more structure than the bare minimum described so far. In particular, each of the sets **F** and **S** may be hierarchically structured on the basis of either some natural classification scheme or by a scheme induced on information-theoretic grounds. There may also be internal causal connections. This in turn may induce structure on the relation **C**, over and above any uncertainty considerations.

## 2.3. Dealing with Uncertainty

A structual consideration of a different nature is uncertainty, which comes into play most obviously by considering **C** to be a weighted connection. More formally, we can introduce two related functions $C_{F|S}$: **F x S --> W** and $C_{S|F}$: **S x F --> W**, with the connotation that $C_{F|S}(f|s)$ is the conditional probability of **f**, given **s**, and $C_{S|F}(s|f)$ is the conditional probability of **s**, given **f**, where **W** is the weight space, say, the unit interval **[0,1]**. Although this is suggestive of a Bayesian paradigm, which is our preferred point of departure, the model is by no means limited to that.

If the sets **F** and/or **S** have additional hierarchical structure, the corresponding connections may be weighted as well, and in the case of causal connections they usually will be. Thus, in general, weights may be attached to any and all arcs in whatever graph structure we have.

Within this setting, uncertainty enters into the picture in three major forms: as a measure of fit, as a discriminant for optimal choice of tests, and as a clustering tool.

**2.3.1. Measure of Fit** The most obvious role of uncertainty is in determining a measure of fit between the input symptom set and the associated faults or fault sets as their explanations. These may be information-theoretic measures, such as, for example, in statistical decision theory or pattern recognition, or they may be heuristic variants of these.

**2.3.2. Optimal Choice of Tests** On the same basis one can compute the discriminating power of potential tests for the presence of additional symptoms, and thus optimize the diagnostic procedure, or at least significantly improve it. This also can be combined with a more heuristic hypothesize-and-test paradigm.

**2.3.3. Clustering** In the multiple fault case the above measures entail clustering techniques, such as found in pattern recognition, which can also be used to structure the fault and symptom sets **F** and **S** in both the single and multiple fault cases.

## 3. Summary and Conclusions

We have presented a model for decision under uncertainty in diagnosis, with emphasis on uncertainty, in support of our advocating the need for a strong underlying model and an integrated set of support tools for carrying such a model in order to deal with uncertainty. Our implementation of this is based on a general purpose rule-based language [4] extended by a comprehensive set of knowledge acquisition tools within which is embedded support for uncertainty in the form presented here, including tools for data and model analysis. We conclude with a summary of our position on the choice of uncertainty models, the propagation of uncertainty in networks, and the acquisition of uncertainty values as three major areas in need of development for the success of our implementation.

### 3.1. Choice of Uncertainty Models

Although our terminology is suggestive of a Bayesian approach, or at least a probabilistic approach, actually there is no commitment to a particular uncertainty model, and we are prepared to support any one of the models, Bayesian, Dempster/Shafer, Fuzzy, and Heuristic, prominent in the AI literature [8]. Our own prefernce is for Bayesian as a point of departure, but any model with a well developed theory will serve, with tests in the field determining the most suitable model for a particular application.

In this context, Heuristic is not so much a model as a practical contingency to which any one of the above models is apt to lead, in that seldom does a problem fit exactly the theoretical assumptions of the background model, and when it does, lack of data to fix the parameters or lack of computing power to use the full model will lead to heuristic approximations. However, it is important to have some background formalism to facilitate the development as well as the analysis of such heuristic models.

### 3.2. Propagation of Uncertainty in Networks

One major problem is the propensity of AI models to give rise to complex networks along which the uncertainties must be propagated and updated. In our model this comes into the picture in two places. First, the hierarchical structuring will lead to such networks. Second, as in any rule-based sytem, there is an implicit inference network associated with the rule set.

This requires both the development of correct propagation models and the translation of them to viable algorithms, since computational complexity comes into the picture as well. However, although theoretically the problems are always NP-Complete when pushed to full generality, this really is not that serious an impediment in practice, since one can generally settle on viable heuristic approximations, and fixed problems do not grow to infinity. Some practical algorithms along these lines are described in [6, 3] and references cited therein.

### 3.3. Acquisition of Uncertainty Values

Another major problem is acquisition of the necessary uncertainty values, or for that matter the related problem of fixing on a useful model. For data rich areas there are the standard statistical tools, but for the more traditional AI approach of interviewing experts it is a recognized problem, although it has been attacked with some success [1]. In addition to developing a strong background model and availing ourselves of standard statistical tools, we expect to look for help from experimental psychology , which has faced similar problems quite successfully, a theme also advocated in [3].

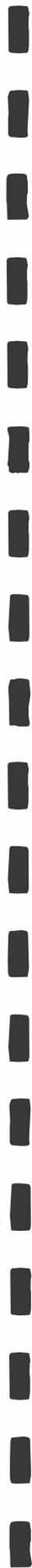